\documentclass[letterpaper]{article} 
\usepackage[preprint]{aaai2027}  
\usepackage[hyphens]{url}  
\usepackage{graphicx} 
\usepackage{natbib}  
\usepackage{caption} 
\usepackage{algorithm}
\usepackage{algorithmic}

\usepackage{array}
\usepackage{makecell}
\usepackage{multirow}
\usepackage{amsmath,amssymb}
\usepackage{enumitem}
\newcommand{\best}[1]{{\textbf{#1}}}
\newcommand{\second}[1]{{\underline{#1}}}
\usepackage{newfloat}
\usepackage{listings}
\DeclareCaptionStyle{ruled}{labelfont=normalfont,labelsep=colon,strut=off} 
\floatstyle{ruled}
\newfloat{listing}{tb}{lst}{}
\floatname{listing}{Listing}

\usepackage{booktabs}

\title{POEM: Phase-Aware $\mathrm{SO}(2)$ Feature Rotation for Time Series Forecasting Under Periodicity Drift}
\author{
    Jiawen Zhu\textsuperscript{\rm 1},
    Shuhan Liu\textsuperscript{\rm 2},
    Shengxuan Li\textsuperscript{\rm 2},
    Qiming Shi\textsuperscript{\rm 2},
    Di Weng\textsuperscript{\rm 1}\corresponding
}

\affiliations{
    \textsuperscript{\rm 1}School of Software Technology, Zhejiang University\\
    \textsuperscript{\rm 2}State Key Lab of CAD\&CG, Zhejiang University
}

\begin{document}

\maketitle

\begin{abstract}
Deep learning has advanced time series forecasting, but periodicity drift, in which cycle timing and phase vary over time, remains a challenging problem.
Existing methods predominantly model these sequences on fixed time grids, suffering from a limited ability to accommodate phase-related variation.
To address this limitation, we propose \textbf{POEM}, a phase-aware forecasting framework based on latent feature rotation using the special orthogonal group in two dimensions, denoted by $\mathrm{SO}(2)$.
POEM aims to reduce the phase-related variability by learning a phase-correction coordinate and applying an invertible $\mathrm{SO}(2)$-based rotation to paired latent features.
To extrapolate this correction coordinate, Directional Phase Increment Attention (DPIA) retrieves historical phase increments from similar temporal contexts and integrates them into future phase corrections.
Experiments demonstrate that POEM achieves competitive performance, while qualitative visualizations suggest that the learned phase-aware transformation makes latent trajectories more regular.
\end{abstract}

\section{Introduction}
Time series forecasting is essential for applications, ranging from intelligent traffic management~\cite{lv2025ripcn, Shao2026HyperD} and weather forecasting~\cite{tian2026arrow, huang2026storm} to financial~\cite{koa2026reasoning} and energy planning~\cite{Tian2025Energy}.
In recent years, a growing number of deep learning methods~\cite{wu2023timesnet, fan2022depts, Zhang2025MLF} have pushed the boundaries of forecasting performance by extracting complex temporal dependencies from massive historical observations.
Much of this success comes from their ability to identify cyclic regularities and extrapolate them forward.
Yet this strategy rests on an idealized assumption that is rarely stated: that such regularities recur at the same rate and with the same timing as in the past.

\begin{figure}[t]
  \begin{center}
  \includegraphics[width=\linewidth]{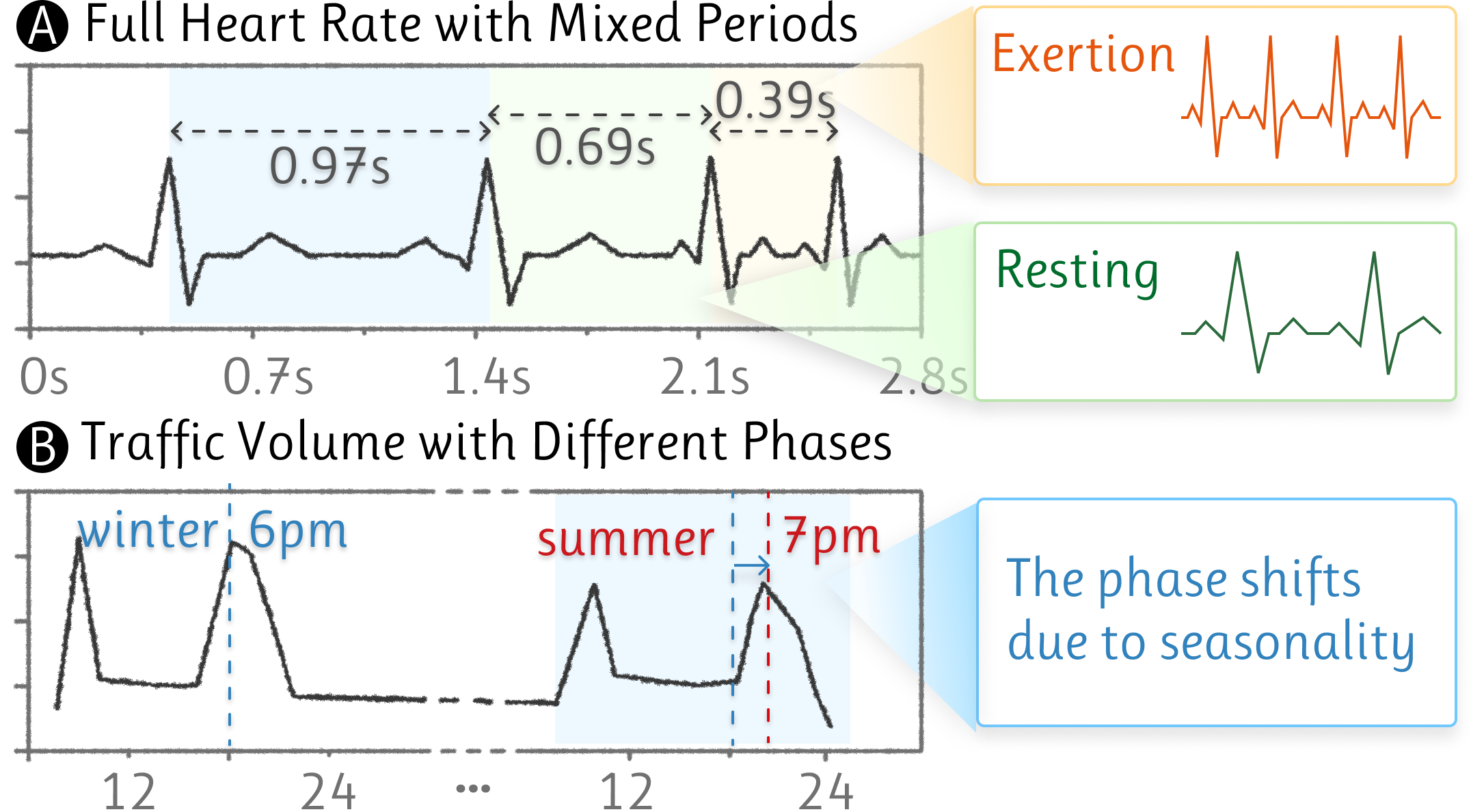}
  \caption{
    \textbf{Two examples of periodicity drift.}
    Cyclic patterns deviate from strict periodicity in real-world data.
    (A) Heart-rate cycles compress under exertion and lengthen in deep sleep.
    (B) The timing and width of a city's daily traffic peaks vary across seasons, with the evening peak occurring slightly later in summer.
    }
  \label{fig:example}
  \end{center}
\end{figure}

Real-world time series, however, rarely repeat themselves so faithfully.
Cycles stretch and compress as operating conditions change, and recurrent events arrive earlier or later than before.
A cardiac cycle of roughly one second at rest compresses under exertion and lengthens during deep sleep, changing the rate at which the cycle advances;
likewise, the timing and width of daily traffic peaks can vary systematically across seasons, with the evening peak occurring slightly later in summer (Figure~\ref{fig:example}).

We refer to these time-varying deviations as \emph{periodicity drift}, and describe them as non-uniform phase progression relative to the uniformly sampled observation clock:
a local phase velocity governs how fast the cycle advances, and a phase offset governs where it is anchored.
Such drift can weaken the correspondence between historical and future cycles, making forecasting more challenging.

Existing methods that model periodicity explicitly fall into two families.
Template-based approaches~\cite{kou2025cfpt, lin2024cyclenet, lin2024sparsetsf}, which model recurrent temporal patterns as forecasting templates, provide a useful inductive bias when cycles are relatively stable, but are less flexible when progression rates and timing vary locally.
Phase-aware methods, such as AlignTime~\cite{Wang2026AlignTime} and PhaseFormer~\cite{niu2026phaseformer}, relax this assumption through dominant period alignment or discrete phase windows.
These approaches account for phase variability, but their global or discretized alignment mechanisms do not explicitly construct a channel-wise phase trajectory from local evidence.
This leaves room for a method that learns phase progression along the observed sequence and extends it coherently into forecast horizon.

To this end, we propose \textbf{POEM}, a phase-aware forecasting framework designed to mitigate the effect of periodicity drift.
Our key insight is that rotating paired features according to the estimated phase-correction trajectory can compensate for phase shifts, yielding more regular temporal patterns and simplifying the subsequent forecasting task.
Concretely, POEM first estimates a phase offset and phase velocity for each channel and local window, forming a phase-correction coordinate.
Second, it applies a rotation from the special orthogonal group in two dimensions, denoted by $\mathrm{SO}(2)$, to reduce phase-related temporal variation, projecting the physical latent states onto a more regular periodic orbit where a lightweight MLP can effectively forecast future states.
Third, Directional Phase Increment Attention (DPIA) retrieves historical phase increments observed under similar timestamp contexts to construct the future phase-correction trajectory.
Finally, the predicted trajectory is applied with the opposite rotation to restore the future phase variation.

Our main contributions are summarized as follows:
\begin{itemize}
    \item
    We \textbf{formulate periodicity drift}, identifying it as a coupling between a system's intrinsic dynamics and a time-varying phase that distorts the observed time axis.
    \item
    We propose \textbf{POEM}, which constructs a continuous phase correction trajectory and applies $\mathrm{SO}(2)$ transformations to features affected by phase shifts, producing a more regular representation for a lightweight temporal predictor.
    \item
    Extensive experiments on real-world benchmarks demonstrate that POEM achieves competitive forecasting performance, while ablation studies and visualization support the effectiveness of POEM.
\end{itemize}

\section{Related Work}
\subsection{Periodic and Phase-Aware Modeling}
Recent forecasting methods increasingly exploit explicit periodic structure to improve performance.
TimesNet~\cite{wu2023timesnet} constructs representations in two dimensions based on the fixed period, while PDF~\cite{dai2024pdf} separates the sequence using multiple period lengths to capture variations over short and long temporal ranges.
Lightweight architectures such as SparseTSF~\cite{lin2024sparsetsf} and CycleNet~\cite{lin2024cyclenet} also explicitly model recurrent cycles with fixed length. 
However, these methods typically rely on fixed-length templates and cannot deform continuously over time.

More recent studies model phase variability more directly.
AlignTime~\cite{Wang2026AlignTime} computes a globally dominant period and aggregates subsequences aligned by phase, while PhaseFormer~\cite{niu2026phaseformer} performs predictions by phase using compact phase embeddings.
PHAT~\cite{ma2026phat} organizes variables into periodic buckets, and TimeAPN~\cite{hu2026timeapn} models phase discrepancies in the frequency domain.
Although these methods account for phase variability, they mainly model it through alignment based on stable global periods, without explicitly representing its evolution along the observed timeline.
In contrast, POEM treats periodicity drift as a temporally evolving phase process and incorporates it into feature learning through a unified geometric framework.

\subsection{Latent Dynamics Forecasting and Geometric Canonicalization}
Several forecasting methods view time series as observations of an underlying dynamical process and seek to model how its state evolves.
Koopa~\cite{liu2023koopa} learns Koopman embeddings to separate time-variant and time-invariant dynamics.
Neural Lad~\cite{li2023neural} models latent evolution with an ODE-based framework enhanced by seasonality-trend characterization, and Attraos~\cite{hu2024attractor} reconstructs phase space and memorizes attractor structures from a chaos perspective.
DeepEDM~\cite{majeedi2025deepedm} and LatentTSF~\cite{yang2026latenttsf} further learn structured latent states instead of regressing in observation space.

These works focus on learning general state evolution or structured representations of temporal dynamics.
POEM shares this dynamical systems perspective but addresses a more specific problem: the nonuniform phase progression associated with periodicity drift.
Rather than aiming to recover the complete underlying dynamics, POEM models the effect of this phase process through structured $\mathrm{SO}(2)$ transformations in feature space.

\section{Methods}
\subsection{Problem Formulation}
\paragraph{Time Series Forecasting}
Let $\mathcal{X} = \{\mathbf{x}_1, \mathbf{x}_2, \dots, \mathbf{x}_L\} \in \mathbb{R}^{L \times C}$ be a multivariate time series observation over a look-back window $L$, where $C$ denotes the number of variables.
The standard time series forecasting task aims to learn a mapping function $\mathcal{F}$ to predict the future sequence $\mathcal{Y} = \{\mathbf{x}_{L+1}, \dots, \mathbf{x}_{L+H}\} \in \mathbb{R}^{H \times C}$ over a horizon $H$, formulated as $\mathcal{Y} = \mathcal{F}(\mathcal{X})$.
From the perspective of non-linear dynamical systems, the discrete observation $\mathbf{x}_t$ can be viewed as a uniform sampling of an observation map from a continuous underlying state $\mathbf{z}(t)$ residing in a phase space, i.e., $\mathbf{x}_t = h(\mathbf{z}(t))|_{t=t_k}$, with $t_k = k\Delta t$~\cite{Takens1981}.

\paragraph{Periodicity Drift}
Many forecasting models effectively rely on relatively stable recurrent patterns, i.e., the system evolves at a constant base frequency $\omega_0$.
Under this idealization, the phase grows linearly with time: $\theta_{ideal}(t) = \omega_0 t$.
In practice, environmental perturbations in real world cause the instantaneous frequency to fluctuate around the base frequency: $\omega(t) = \omega_0 + \delta\omega(t)$, where $\delta\omega(t)$ denotes a small time-varying perturbation.
Consequently, the true instantaneous phase $\varphi(t)$ exhibits a non-linear accumulation:
\begin{equation}
    \varphi(t) = \int_{0}^{t} \omega(\tau) d\tau = \omega_0 t + \Delta \Phi(t)
\end{equation}
where $\Delta\Phi(t)=\int_{0}^{t}\delta\omega(\tau)\,d\tau$ denotes the phase drift.
\begin{figure*}[ht]
  \begin{center}
  \includegraphics[width=\textwidth]{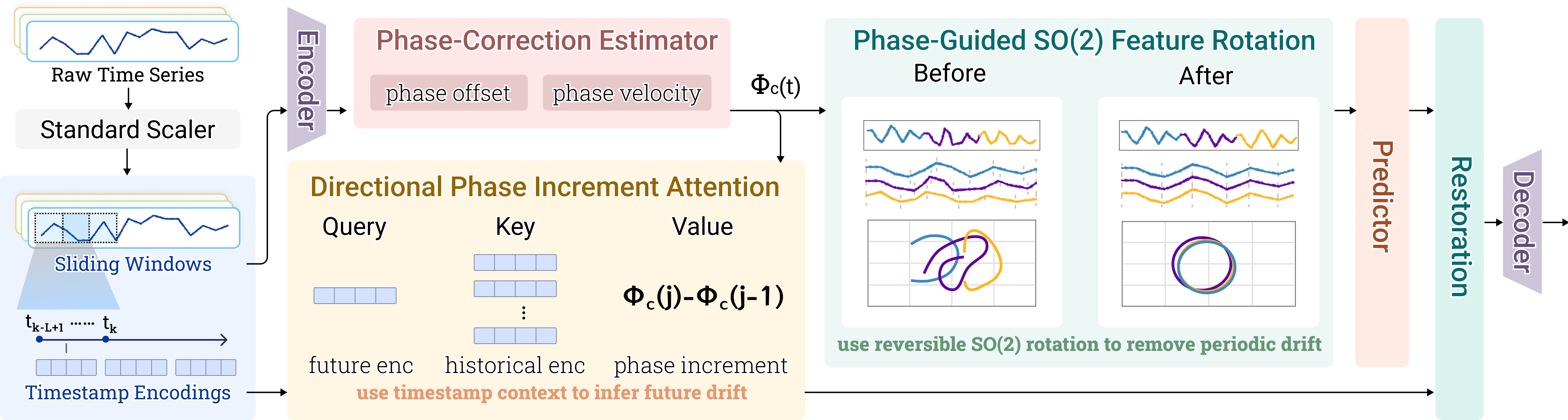}
  \caption{
      Overview of the POEM framework.
      The framework comprises five modules: the Local Phase-Correction Estimator, Phase-Guided $\mathrm{SO}(2)$ Feature Rotation, Phase-Corrected Temporal Predictor, Directional Phase Increment Attention, and Phase Restoration and Decoding.
  }
  \label{fig:overview}
  \end{center}
\end{figure*}

\subsection{Overview}
POEM addresses periodicity drift by estimating variable-specific phase correction trajectories and using them to reduce phase-related variation before forecasting.

The overall framework consists of the following five modules (Figure~\ref{fig:overview}):
\textbf{Local Phase-Correction Estimator (LPCE)} first extracts local phase offsets and velocities for each variable and integrates the overlapping window estimates into a continuous phase correction trajectory.
Guided by this trajectory, \textbf{Phase-Guided $\mathrm{SO}(2)$ Feature Rotation} transforms the paired features into a more regular representation, after which the \textbf{Phase-Corrected Temporal Predictor} forecasts future evolution with a lightweight MLP.
Because phase variation continues into the forecast horizon, \textbf{Directional Phase Increment Attention} retrieves historical phase increments from similar timestamp contexts and uses them to construct the future phase correction trajectory.
Finally, \textbf{Phase Restoration and Decoding} restores the predicted phase variation and produces the final forecast $\widehat{\mathbf{Y}}$.

\subsection{Local Phase-Correction Estimator}
To capture local variations in phase progression, we estimate two complementary quantities for each window $w$ and variable $c$: (1) a phase offset $\theta_{w,c}$, which serves as a window-level angular anchor for the local phase correction; and (2) a positive phase velocity $v_{w,c}$, which describes how quickly the sequence moves through its cycle in the window.
We leverage them together in constructing a phase correction trajectory $\Phi_c(t)$ for each variable.

\paragraph{Local Dynamics Encoder}
We divide the normalized look-back sequence per variable into $W=\lfloor(L-\ell)/s\rfloor+1$ overlapping windows of length $\ell$ and stride $s$, where $\mathbf{X}_{w,c}\in\mathbb{R}^{\ell\times1}$ denotes the $w$-th window of variable $c$.
Each window is processed by an encoder with convolutional mixing, self-attention, and temporal aggregation:
\begin{equation}
\mathbf{z}_{w,c}=\operatorname{Encoder}(\mathbf{X}_{w,c})\in\mathbb{R}^{D}.
\label{eq:local_observation_encoder}
\end{equation}
This resulting window representation $\mathbf{z}_{w,c}$ provides local information for the subsequent phase estimation.

\paragraph{Local Phase Field Parameterization}
Given the representation $\mathbf{z}_{w,c}$, we parameterize two quantities that jointly determine the local phase behavior.

\paragraph{\textit{(i) Phase Offset}}
We predict a local phase offset to construct the window-wise correction:
\begin{equation}
    \theta_{w,c} = \pi\cdot\tanh\!\big(\mathrm{MLP}_{\theta}(\mathbf{z}_{w,c})\big),\quad
    \theta_{w,c}\in(-\pi,\pi).
\end{equation}
Constraining $\theta_{w,c}$ within $[-\pi,\pi]$ respects the $2\pi$-periodicity of the underlying cyclic state and stabilizes optimization.

\paragraph{\textit{(ii) Phase Velocity}} 
We parameterize a positive local phase velocity for each channel and window, ensuring that the window-local intrinsic clock advances forward.
We enforce this by predicting its logarithm via a clipped MLP and exponentiating:
\begin{equation}
    \rho_{w,c} = \mathrm{Clip} \big(\mathrm{MLP}_{v}(\mathbf{z}_{w,c}),\,[-\gamma,\,\gamma]\big)
\end{equation}
\begin{equation}
    v_{w,c} = \exp(\rho_{w,c})
\end{equation}
where the clipping threshold $\gamma$ bounds the admissible range of temporal distortion.
Integrating the dense velocity field over the stride $s$ yields the accumulated phase time $\tau_{w,c}$ at the starting position of window $w$:
\begin{equation}
    \tau_{1,c} = 0, \quad
    \tau_{w,c} = \sum_{j=1}^{w-1} v_{j,c}s, \quad 
    w=2,\ldots,W.
\end{equation}
which represents how much intrinsic phase progress has elapsed by that physical step.

\paragraph{Global Continuous Phase Integrator}
The window-level estimation is integrated into a coherent phase trajectory.
Let $t_w=(w-1)s$ denote the starting index of window $w$.
For channel $c$, we propagate the local phase anchor $\theta_{w,c}$ within each window according to its phase velocity:
\begin{equation}
\psi_{w,c}(t)
  = \theta_{w,c}
   + \frac{2\pi}{l_p}
     \left[\tau_{w,c}+v_{w,c}(t-t_w)-t\right].
\label{eq:window_phase_propagation}
\end{equation} 
Here, $\tau_{w,c}$ is the accumulated intrinsic phase time at the window start, $l_p$ denotes the global dominant period which is predefined at the dataset level, and $2\pi/l_p$ converts the deviation from the physical clock into an angular correction.
We then aggregate them by a circular mean and temporally unwrap the result:
\begin{equation}
\begin{aligned}
\widetilde{\Phi}_c(t)
  &= \operatorname{atan2}\!\left(
     \sum_{w\in\mathcal W(t)}\sin\psi_{w,c}(t),
     \sum_{w\in\mathcal W(t)}\cos\psi_{w,c}(t)
     \right), \\
\Phi_c(t)
  &= \begin{cases}
     \widetilde{\Phi}_c(1), & t=1, \\[2pt]
     \Phi_c(t-1)+\delta_c(t), & 2\leq t\leq L.
     \end{cases}
\end{aligned}
\label{eq:circular_phase_fusion}
\end{equation}
Here, $\delta_c(t)=\operatorname{wrap}_{\pi}(\widetilde{\Phi}_c(t)-\widetilde{\Phi}_c(t-1))$, where $\operatorname{wrap}_{\pi}(a)=\operatorname{atan2}(\sin a,\cos a)\in[-\pi,\pi]$.
This produces a continuous, channel-wise phase correction trajectory $\Phi_c(t)$, which derives the subsequent $\mathrm{SO}(2)$ latent rotation.

\subsection{Phase-Guided $\mathrm{SO}(2)$ Feature Rotation}
\begin{figure}[t]
  \begin{center}
  \includegraphics[width=\linewidth]{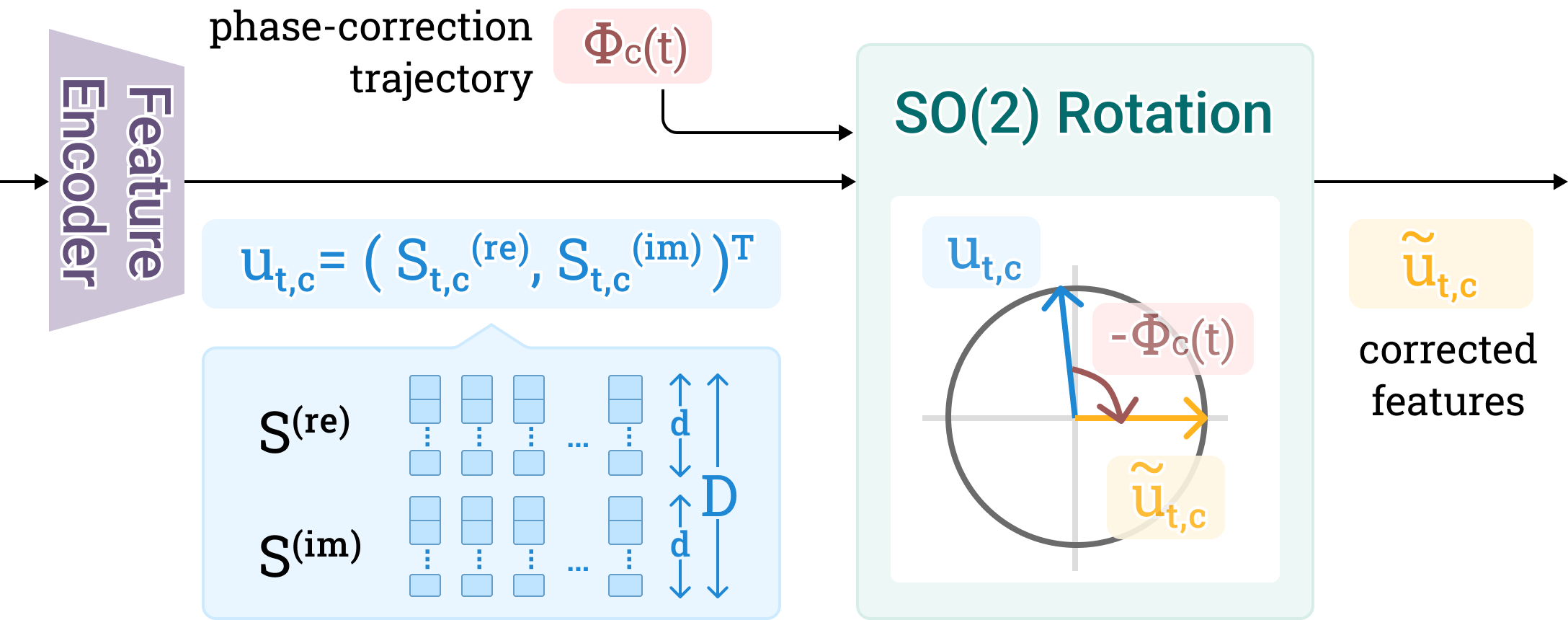}
  \caption{
     Phase-guided $\mathrm{SO}(2)$ feature rotation.
    }
  \label{fig:rotation}
  \end{center}
\end{figure}

We represent phase transformations using the planar rotation group $\mathrm{SO}(2)$, as shown in Figure~\ref{fig:rotation}.
For an angle $\phi$, its action on a two-dimensional latent feature pair $\mathbf{u}\in\mathbb{R}^{2}$ is given by
\begin{equation}
R(\phi)=
\begin{bmatrix}
\cos\phi & -\sin\phi \\
\sin\phi & \cos\phi
\end{bmatrix}.
\label{eq:so2_rotation}
\end{equation}
This transformation is invertible since $R(\phi)^{\top}R(\phi)=I$ and $R(\phi)^{-1}=R(-\phi)$.

\paragraph{Temporal Feature Encoding}
Before the phase-guided rotation, we encode the normalized observations $\bar{\mathcal{X}}$ into a $D$-dimensional feature per variable and time step.
A linear projection first maps each observation to $D$-dimensional vector, after which a temporal convolution and a token-mixing MLP model local patterns and dependencies over the look-back sequence.
Collecting the encoded features gives
\begin{equation}
\mathbf{S}
=\operatorname{FeatureEncoder}(\bar{\mathcal{X}})
\in\mathbb{R}^{L\times C\times D}.
\label{eq:temporal_feature_encoding}
\end{equation}
$\mathbf{s}_{t,c}=\mathbf{S}_{t,c,:}\in\mathbb{R}^{D}$ is the feature of variable $c$ at time step $t$.

\paragraph{$\mathrm{SO}(2)$ Rotation on Paired Features}
The encoded features still retain the phase shifts.
We reduce this effect by rotating the feature vector by the negative phase-correction angle $-\Phi_c(t)$.
Let $D=2d$, and split each feature vector into two blocks $\mathbf{s}_{t,c}^{(\mathrm{re})},\mathbf{s}_{t,c}^{(\mathrm{im})}\in\mathbb{R}^{d}$ whose corresponding coordinates form $d$ two-dimensional feature pairs.
The variable-specific phase correction $\Phi_c(t)$ is applied to every pair through
\begin{equation}
\begin{bmatrix}
\widetilde{s}_{t,c,k}^{(\mathrm{re})} \\
\widetilde{s}_{t,c,k}^{(\mathrm{im})}
\end{bmatrix}
=R\!\left(-\Phi_c(t)\right)
\begin{bmatrix}
s_{t,c,k}^{(\mathrm{re})} \\
s_{t,c,k}^{(\mathrm{im})}
\end{bmatrix},
\quad k=1,\ldots,d.
\label{eq:phase_canonicalization}
\end{equation}
Rotating by $-\Phi_c(t)$ can compensate for the estimated phase shift, making the transformed features more regular and easier for the subsequent predictor to forecast.

Concatenating the rotated blocks gives $\widetilde{\mathbf{s}}_{t,c}=[\widetilde{\mathbf{s}}_{t,c}^{(\mathrm{re})};\widetilde{\mathbf{s}}_{t,c}^{(\mathrm{im})}]\in\mathbb{R}^{D}$, and collecting them over time and variables yields $\widetilde{\mathbf{S}}\in\mathbb{R}^{L\times C\times D}$.

\subsection{Phase-Corrected Temporal Predictor}
The $\mathrm{SO}(2)$ rotation reduces phase-induced variation in $\widetilde{\mathbf{S}}$, allowing a lightweight predictor to focus on the remaining temporal dependencies.
We apply MLP blocks along the look-back dimension while keeping the variable and feature dimensions unchanged.
A linear horizon projection then maps the temporal dimension from $L$ to $H$, producing the future phase-corrected features $\widehat{\mathbf{S}}_{\mathrm{pc}}\in\mathbb{R}^{H\times C\times D}$.

\subsection{Directional Phase Increment Attention}
Although the predictor operates on phase-corrected features, the future observations are still expected to exhibit periodicity drift.
We therefore forecast a future phase-correction trajectory and use it to rotate the predicted features back.
We construct this trajectory by retrieving historical phase increments observed under similar temporal contexts.

We implement this with Directional Phase Increment Attention (DPIA), which compares normalized temporal-context embeddings via a learnable scaled cosine kernel.

\paragraph{Directional Queries and Keys}
Let $\mathbf{M}_{\mathrm{enc}}\in\mathbb{R}^{L\times d_m}$ and $\mathbf{M}_{\mathrm{dec}}\in\mathbb{R}^{H\times d_m}$ denote the timestamp encodings associated with the historical and future horizons, respectively.
The future timestamp encodings serve as queries, while historical timestamp encodings at $j=2,\ldots,L$ serve as keys so that each key is aligned with the phase increment ending at the same time step.
Using learnable projections $W_Q,W_K\in\mathbb{R}^{d_m\times d_m}$, we define
\begin{equation}
\begin{aligned}
\mathbf{q}_h
&=\frac{\mathbf{m}^{\mathrm{dec}}_hW_Q}
{\lVert\mathbf{m}^{\mathrm{dec}}_hW_Q\rVert_2},
&&h=1,\ldots,H,\\
\mathbf{k}_j
&=\frac{\mathbf{m}^{\mathrm{enc}}_jW_K}
{\lVert\mathbf{m}^{\mathrm{enc}}_jW_K\rVert_2},
&&j=2,\ldots,L.
\end{aligned}
\label{eq:directional_query_key}
\end{equation}
Stacking these vectors gives $\mathbf{Q}\in\mathbb{R}^{H\times d_m}$ and $\mathbf{K}\in\mathbb{R}^{(L-1)\times d_m}$.

\paragraph{Channel-wise Phase Increments as Values}
An absolute phase coordinate depends on its angular origin, whereas first differences are invariant to a constant phase offset and directly describe local phase changes.
For each historical time step $j=2,\ldots,L$, we therefore define
\begin{equation}
\begin{aligned}
\mathbf{v}_j
&=\left[\Phi_c(j)-\Phi_c(j-1)\right]_{c=1}^{C}\in\mathbb{R}^{C},
&&j=2,\ldots,L,\\
\mathbf{V}
&=\begin{bmatrix}
\mathbf{v}_2^{\top} & \cdots & \mathbf{v}_L^{\top}
\end{bmatrix}^{\top}
\in\mathbb{R}^{(L-1)\times C}.
\end{aligned}
\label{eq:phase_increment_values}
\end{equation}

\paragraph{Scaled Cosine Attention}
Because the queries and keys are $\ell_2$-normalized, their inner products measure cosine similarity between the projected future and historical timestamp encodings.
The attention weights are
\begin{equation}
\mathbf{A}
=\operatorname{Softmax}\!\left(\kappa\mathbf{Q}\mathbf{K}^{\top}\right)
\in\mathbb{R}^{H\times(L-1)},
\label{eq:directional_attention}
\end{equation}
where the softmax is applied over historical time steps.
We parameterize the positive scale as
\begin{equation}
\kappa=\operatorname{Softplus}(\kappa_{\mathrm{raw}})+\epsilon.
\label{eq:attention_scale}
\end{equation}
Here, $\kappa_{\mathrm{raw}}\in\mathbb{R}$ is a learnable scalar, and $\epsilon>0$ is a small constant for numerical stability.
A larger $\kappa$ concentrates the weights on the most similar historical timestamps, whereas a smaller value produces smoother aggregation.

\begin{table*}[t]
\centering

{
\fontsize{9}{10.5}\selectfont
\rmfamily
\setlength{\tabcolsep}{0.95mm}
\renewcommand{\arraystretch}{1.05}

\begin{tabular}{@{}l*{9}{cc}@{}}
\toprule
\multirow{2}{*}{\textbf{Dataset}}
  & \multicolumn{2}{c}{\shortstack{\textbf{POEM}\\\textbf{(Ours)}}}
  & \multicolumn{2}{c}{\shortstack{PMDformer\\(2026)}}
  & \multicolumn{2}{c}{\shortstack{Phaseformer\\(2026)}}
  & \multicolumn{2}{c}{\shortstack{SRSNet\\(2025)}}
  & \multicolumn{2}{c}{\shortstack{TQNet\\(2025)}}
  & \multicolumn{2}{c}{\shortstack{TimeBridge\\(2025)}}
  & \multicolumn{2}{c}{\shortstack{CycleNet\\(2024)}}
  & \multicolumn{2}{c}{\shortstack{TimeMixer\\(2024)}}
  & \multicolumn{2}{c}{\shortstack{DLinear\\(2023)}} \\

\cmidrule(lr){2-3}
\cmidrule(lr){4-5}
\cmidrule(lr){6-7}
\cmidrule(lr){8-9}
\cmidrule(lr){10-11}
\cmidrule(lr){12-13}
\cmidrule(lr){14-15}
\cmidrule(lr){16-17}
\cmidrule(lr){18-19}

  & MSE & MAE
  & MSE & MAE
  & MSE & MAE
  & MSE & MAE
  & MSE & MAE
  & MSE & MAE
  & MSE & MAE
  & MSE & MAE
  & MSE & MAE \\
\midrule

Weather
  & \best{0.240} & 0.271
  & \best{0.240} & \best{0.267}
  & 0.262 & 0.282
  & 0.260 & 0.282
  & \second{0.242} & \second{0.269}
  & 0.254 & 0.270
  & 0.254 & 0.279
  & 0.246 & 0.275
  & 0.268 & 0.318 \\

Exchange
  & \second{0.349} & \best{0.396}
  & 0.583 & 0.474
  & 0.363 & 0.409
  & 0.361 & 0.403
  & 0.369 & 0.407
  & 0.395 & 0.426
  & 0.373 & 0.407
  & 0.395 & 0.418
  & \best{0.336} & \second{0.402} \\

ETTh1
  & \best{0.422} & 0.431
  & 0.442 & 0.436
  & \second{0.434} & \second{0.430}
  & 0.442 & 0.434
  & 0.447 & 0.440
  & 0.446 & 0.440
  & \second{0.434} & \best{0.428}
  & 0.455 & 0.443
  & 0.459 & 0.452 \\

ETTh2
  & \best{0.375} & 0.401
  & 0.382 & 0.401
  & 0.382 & 0.410
  & \second{0.376} & \second{0.400}
  & 0.378 & 0.402
  & 0.379 & \best{0.399}
  & 0.385 & 0.404
  & 0.378 & 0.405
  & 0.498 & 0.479 \\

ETTm1
  & \best{0.371} & \second{0.393}
  & 0.403 & 0.395
  & 0.399 & 0.401
  & 0.390 & 0.398
  & 0.386 & 0.399
  & \second{0.380} & \best{0.389}
  & 0.387 & 0.395
  & 0.384 & 0.398
  & 0.407 & 0.410 \\

ETTm2
  & \best{0.274} & 0.322
  & 0.280 & \second{0.319}
  & 0.290 & 0.334
  & 0.284 & 0.330
  & 0.278 & 0.322
  & 0.282 & 0.322
  & \second{0.275} & \best{0.316}
  & 0.279 & 0.325
  & 0.310 & 0.367 \\

NN5
  & \second{0.741} & \second{0.602}
  & 0.823 & 0.654
  & 0.971 & 0.732
  & 0.808 & 0.648
  & 0.765 & 0.619
  & \best{0.731} & \best{0.592}
  & 0.763 & 0.620
  & 0.787 & 0.634
  & 1.184 & 0.828 \\

US Birth
  & \second{0.301} & \second{0.386}
  & 0.580 & 0.598
  & 0.574 & 0.545
  & 0.370 & 0.452
  & 0.336 & 0.418
  & 0.358 & 0.424
  & 0.369 & 0.450
  & \best{0.233} & \best{0.344}
  & 0.502 & 0.540 \\

ZafNoo
  & 0.537 & 0.452
  & \best{0.499} & \best{0.426}
  & 0.583 & 0.465
  & 0.593 & 0.479
  & 0.563 & 0.465
  & 0.558 & \second{0.446}
  & 0.593 & 0.480
  & 0.556 & 0.461
  & \second{0.518} & 0.466 \\
\midrule

Top-2 count
  & \multicolumn{2}{c}{\best{12}}
  & \multicolumn{2}{c}{5}
  & \multicolumn{2}{c}{2}
  & \multicolumn{2}{c}{2}
  & \multicolumn{2}{c}{2}
  & \multicolumn{2}{c}{6}
  & \multicolumn{2}{c}{4}
  & \multicolumn{2}{c}{2}
  & \multicolumn{2}{c}{3} \\
\bottomrule
\end{tabular}
}

\caption{Time series forecasting performance. Results are averaged across four
prediction horizons: $\{24,36,48,60\}$ for the NN5 benchmark and
$\{96,192,336,720\}$ for the remaining datasets. The look-back window length
is set to 36 for NN5 and 96 for the other datasets. Lower MSE and MAE values
indicate better forecasting accuracy. The best and second-best results are
shown in \best{bold} and \second{underlined}, respectively. Detailed results are provided
in Appendix.}
\label{tab:main_results}

\end{table*}

\paragraph{Future Phase Integration}
The predicted phase increments are obtained by aggregating historical increments with the attention weights:
\begin{equation}
\boldsymbol{\delta}=\mathbf{A}\mathbf{V}\in\mathbb{R}^{H\times C}.
\label{eq:future_phase_increments}
\end{equation}
We recover the future phase-correction trajectory by cumulative summation:
\begin{equation}
\widehat{\Phi}_c(L+h)
=\Phi_c(L)+\sum_{\tau=1}^{h}\delta_{\tau,c},
\quad h=1,\ldots,H.
\label{eq:future_phase_integration}
\end{equation}
The trajectory is used by the subsequent phase restoration rotation to reintroduce the forecast phase variation.

\subsection{Phase Restoration and Decoding}
The temporal predictor produces future features in the phase-corrected representation, whereas the final forecasts should reflect the phase evolution over the prediction horizon.
We thus rotate each predicted feature pair by its corresponding future phase-correction angle.
For $h=1,\ldots,H$, variable $c$, and feature pair $k=1,\ldots,d$, we compute
\begin{equation}
\begin{bmatrix}
\widehat{s}_{h,c,k}^{(\mathrm{re})}\\
\widehat{s}_{h,c,k}^{(\mathrm{im})}
\end{bmatrix}
=R\!\left(\widehat{\Phi}_c(L+h)\right)
\begin{bmatrix}
\widehat{s}_{\mathrm{pc},h,c,k}^{(\mathrm{re})}\\
\widehat{s}_{\mathrm{pc},h,c,k}^{(\mathrm{im})}
\end{bmatrix}.
\label{eq:future_phase_restoration}
\end{equation}
thereby restoring the phase variation in the predicted features.
Collecting the restored feature vectors gives $\widehat{\mathbf{S}}_{\mathrm{rest}}\in\mathbb{R}^{H\times C\times D}$.
Then, a linear decoder $g_{\mathrm{dec}}$ maps the restored features to the forecast:
\begin{equation}
\widehat{\mathbf{Y}}
=g_{\mathrm{dec}}\!\left(\widehat{\mathbf{S}}_{\mathrm{rest}}\right)
\in\mathbb{R}^{H\times C}.
\label{eq:forecast_decoding}
\end{equation}
The forecast is finally returned to the original data scale by reversing the input normalization.

\section{Experiments}
\subsection{Experiment Settings}
\paragraph{Datasets}
We evaluate POEM on nine real-world benchmarks: ETT (four subsets)~\cite{Zhou2021Informer}, Exchange~\cite{Lai2018Exchange}, Weather~\cite{wu2021autoformer}, NN5~\cite{crone2008nn5}, ZafNoo~\cite{poyatos2019zafnoo}, and US Birth~\cite{godahewa2020usbirths}.
These datasets span diverse domains, including energy, finance, and climatology.
The details of each dataset are summarized in Appendix.

\paragraph{Implementation Details}
All experiments are implemented in PyTorch~\cite{paszke2019pytorch} and conducted on a single NVIDIA A100 $80$\,GB GPU.
We use 2026 as the random seed for all experiments.
POEM is optimized with Adam using mean squared error (MSE) as the training objective, and the initial learning rate is selected between $1 \times 10^{-3}$ and $5 \times 10^{-3}$.
The look-back length is fixed to $L=96$ and the forecasting horizons to $H\in\{96,192,336,720\}$, except for NN5, where $L=36$ and $H\in\{24,36,48,60\}$.
Models are trained for at most 50 epochs with early stopping based on validation loss, using a patience of 5 epochs for Weather and 10 epochs for the remaining datasets.
We report MSE and mean absolute error (MAE) as the evaluation metrics.
The hyperparameters for each dataset and forecasting horizon are provided in Appendix.

\subsection{Main Results}
\paragraph{Baselines}
We compare POEM with eight representative baselines: PMDformer~\cite{hu2026pmdformer}, Phaseformer~\cite{niu2026phaseformer}, SRSNet~\cite{wu2026srsnet}, TQNet~\cite{lin2025tqnet}, TimeBridge~\cite{liu2025timebridge}, CycleNet~\cite{lin2024cyclenet}, TimeMixer~\cite{wang2024timemixer}, and Dlinear~\cite{Zeng2023Dlinear}.

\paragraph{Results Analysis}
Table~\ref{tab:main_results} summarizes the MSE and MAE averaged over four prediction horizons for each dataset.
Across the nine datasets and two evaluation metrics, POEM achieves either the best or second-best result in \textbf{12} of the \textbf{18} comparisons, the highest count among all models.
Compared with leading baseline PMDformer, POEM reduces the average MSE by \textbf{11.9\%}, surpassing it on \textbf{7} out of \textbf{9} datasets.
The performance suggests that our method has a significant effect in addressing the inherent challenges of periodicity drift in time series forecasting.

\subsection{Ablation Study}
We evaluate four ablated variants on eight datasets (three shown here, full results in Appendix) to examine the contribution of each component in POEM, with the results reported in Table~\ref{tab:ablation_results}.
We also separately replace the $\mathrm{SO}(2)$ rotation with three alternative phase transformations and report this comparison in Table~\ref{tab:ablation_SO2}.
Full results are provided in the Appendix.

\begin{table}[t]
\centering

{
\fontsize{9}{10.5}\selectfont
\rmfamily
\setlength{\tabcolsep}{1.8pt}
\renewcommand{\arraystretch}{1.1}

\begin{tabular}{@{}l*{3}{cc}@{}}
\toprule

\multirow{2}{*}{\textbf{Variant}}
& \multicolumn{2}{c}{\textbf{ETTh1}}
& \multicolumn{2}{c}{\textbf{ETTm1}}
& \multicolumn{2}{c}{\textbf{US Birth}} \\

\cmidrule(lr){2-3}
\cmidrule(lr){4-5}
\cmidrule(lr){6-7}

& MSE & MAE
& MSE & MAE
& MSE & MAE \\
\midrule

\textbf{POEM}
& \best{0.422} & \best{0.431}
& \best{0.371} & \best{0.393}
& \best{0.301} & \best{0.386} \\

w/o SO(2) Rotation
& 0.451 & 0.444
& 0.380 & 0.399
& 0.399 & 0.449 \\

w/o LPCE
& 0.710 & 0.575
& 0.422 & 0.416
& 1.499 & 1.057 \\

w/o DPIA (Linear)
& 0.457 & 0.445
& 0.378 & 0.397
& 0.435 & 0.483 \\

w/o DPIA (Standard Attn)
& 0.460 & 0.448
& 0.376 & 0.396
& 0.320 & 0.398 \\

\bottomrule
\end{tabular}
}

\caption{
    Ablation results averaged over four horizons on three datasets.
    Detailed results are provided in Appendix.
}
\label{tab:ablation_results}

\end{table}

\paragraph{Effect of Phase-Guided $\mathrm{SO}(2)$ Feature Rotation}
In \textit{w/o $\mathrm{SO}(2)$ Rotation}, the encoded features are passed directly to the temporal predictor and decoder without the phase correction and restoration rotations.
Removing these rotations increases the MSE on every evaluated dataset, including from $0.301$ to $0.399$ on US Birth, and from $0.422$ to $0.451$ on ETTh1.
These results suggest that adjusting the encoded features with $\mathrm{SO}(2)$ feature rotation is beneficial to the subsequent forecasting task.

\paragraph{Effect of Local Phase-Correction Estimator}
In \textit{w/o Local Phase-Correction Estimator (LPCE)}, we remove the local window-based estimation of phase offset and phase velocities, together with the subsequent integration.
Instead, a linear head directly predicts a phase increment from the encoded feature at each time step, and cumulative summation forms the historical phase trajectory.
This variant produces the largest degradation among the four ablations, with the MSE increasing from $0.422$ to $0.710$ on ETTh1 and $0.371$ to $0.422$ on ETTm1.
The consistent degradation across all datasets supports the structured construction of the phase correction trajectory from local offsets and velocities, compared with directly learning per-step increments.

\paragraph{Effect of Directional Phase Increment Attention}
We consider two alternatives to Directional Phase Increment Attention (DPIA).
The \textit{w/o DPIA (Linear Projection)} variant discards the timestamp encodings and maps the complete historical phase-increment sequence directly to future increments through a learned linear projection, while \textit{w/o DPIA (Standard Attention)} replaces the scaled cosine attention with standard scaled dot-product attention.
Both alternatives yield higher MSE and MAE than the complete model on all eight datasets.
For example, the MSE on US Birth increases from $0.301$ to $0.435$ with linear projection and to $0.320$ with standard attention, while its values on ETTh1 increase from $0.422$ to $0.457$ and $0.460$.
These comparisons support retrieving historical phase increments from similar timestamp contexts using the scaled cosine attention in DPIA.

\paragraph{Alternative Transformations to $\mathrm{SO}(2)$ Rotation}
We replace the $\mathrm{SO}(2)$ rotation with three alternative phase transformations while keeping the remaining components unchanged: \textit{Frequency-Domain Phase Correction}, \textit{Phase-Conditioned FiLM}, and \textit{Phase-Conditioned Shear}.
As shown in Table~\ref{tab:ablation_SO2}, the $\mathrm{SO}(2)$ rotation achieves lower MSE and MAE than all three alternatives.
Compared with the strongest alternative, it reduces the MSE from $0.435$ with \textit{Phase-Conditioned FiLM} to $0.422$ on ETTh1 and from $0.349$ with \textit{Phase-Conditioned Shear} to $0.301$ on US Birth.
These results suggest that norm-preserving $\mathrm{SO}(2)$ rotation of paired latent features is more effective than the evaluated frequency-domain, feature-wise affine, and shear transformations.

In all, the ablations support the contribution of the three components to the forecasting performance of POEM, while the comparisons with alternative transformations further support the use of $\mathrm{SO}(2)$ rotation for phase correction.

\begin{table}[t]
\centering

{
\fontsize{9}{10.5}\selectfont
\rmfamily
\setlength{\tabcolsep}{1.8pt}
\renewcommand{\arraystretch}{1.1}

\begin{tabular}{@{}l*{3}{cc}@{}}
\toprule

\multirow{2}{*}{\textbf{Variant}}
& \multicolumn{2}{c}{\textbf{ETTh1}}
& \multicolumn{2}{c}{\textbf{ETTm1}}
& \multicolumn{2}{c}{\textbf{US Birth}} \\

\cmidrule(lr){2-3}
\cmidrule(lr){4-5}
\cmidrule(lr){6-7}

& MSE & MAE
& MSE & MAE
& MSE & MAE \\
\midrule

\textbf{POEM ($\mathrm{SO}(2)$)}
& \textbf{0.422} & \textbf{0.431}
& \textbf{0.371} & \textbf{0.393}
& \textbf{0.301} & \textbf{0.386} \\

Frequency-Domain Phase
& 0.564 & 0.505
& 0.440 & 0.430
& 0.659 & 0.606 \\

Phase-Conditioned FiLM
& 0.435 & 0.440
& 0.381 & 0.398
& 0.460 & 0.500 \\

Phase-Conditioned Shear
& 0.456 & 0.443
& 0.377 & 0.398
& 0.349 & 0.417 \\

\bottomrule
\end{tabular}
}

\caption{
    Comparison with alternative transformations replacing the $\mathrm{SO}(2)$ rotation, averaged over four prediction horizons on three datasets.
    Detailed results are in Appendix.
}
\label{tab:ablation_SO2}

\end{table}

\subsection{Hyperparameter Sensitivity}
\begin{figure}[ht]
  \begin{center}
  \includegraphics[width=\linewidth]{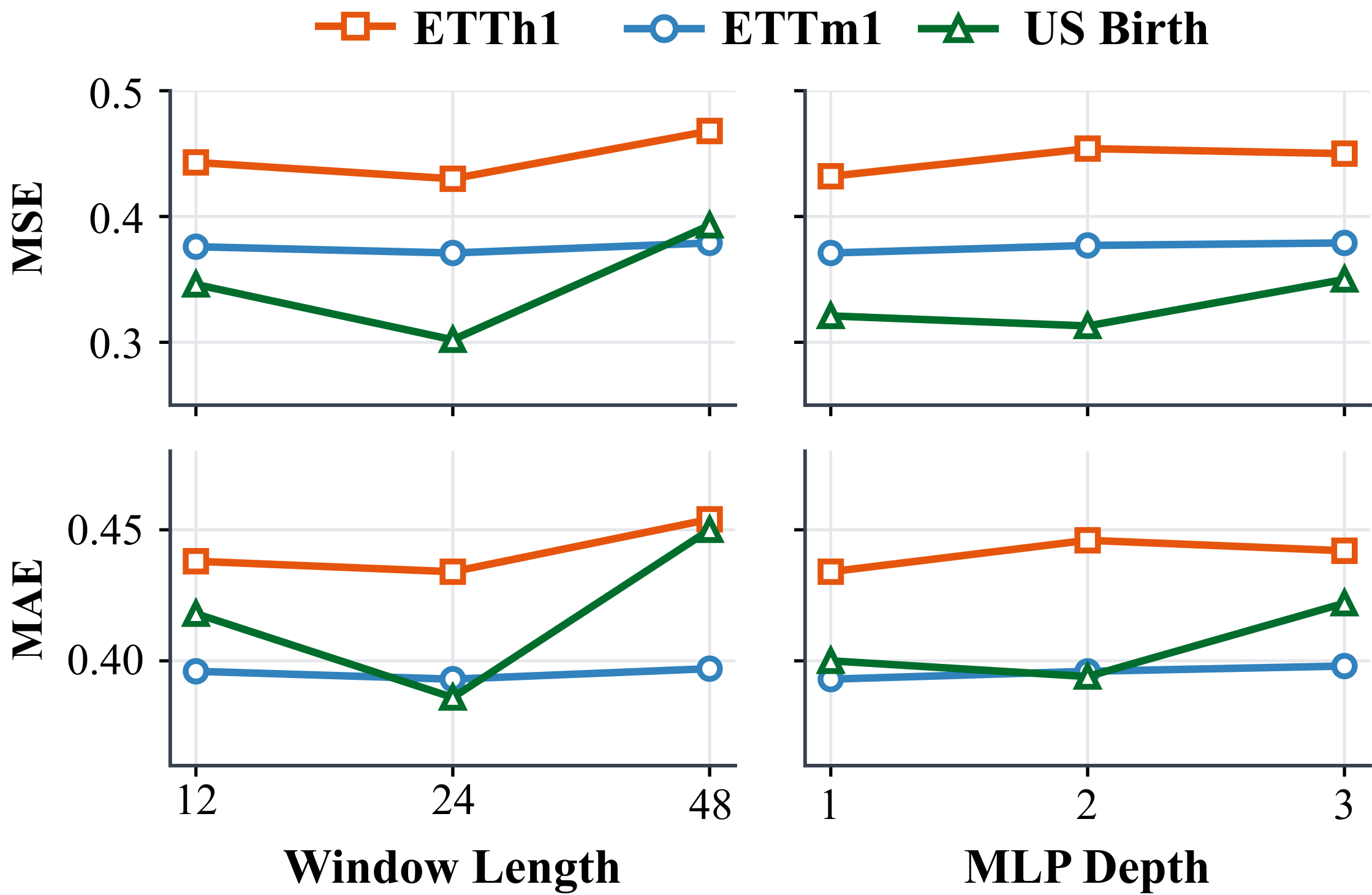}
  \caption{
        Hyperparameter sensitivity of POEM to the local window length $\ell$ and temporal MLP depth $K$ on ETTh1, ETTm1, and US Birth.
  }
  \label{fig:hyperparameter}
  \end{center}
\end{figure}

We examine the sensitivity of POEM to two hyperparameters on ETTh1, ETTm1 and US Birth: the local window length $\ell\in\{12,24,48\}$ and the number of temporal MLP layers $K\in\{1,2,3\}$.
The results are summarized in Figure~\ref{fig:hyperparameter}.

\paragraph{Effect of Window Length $\ell$}
The window length determines the amount of local temporal context used by the Local Phase-Correction Estimator.
Among the tested values, $\ell=24$ achieves the lowest MSE and MAE on all three datasets.
The sensitivity is relatively small on ETTm1, where the MSE ranges from $0.371$ to $0.379$, but is more pronounced on US Birth, where the MSE increases from $0.302$ at $\ell=24$ to $0.346$ at $\ell=12$ and $0.393$ at $\ell=48$.
These results favor a moderate local window in the evaluated settings, although the magnitude of its effect varies across datasets.

\paragraph{Effect of Temporal MLP Depth $K$}
To examine the required predictor capacity, we vary the depth of the temporal MLP predictor from one to three layers.
A single layer achieves the lowest MSE and MAE on ETTh1 and ETTm1, whereas two layers perform best on US Birth.
Increasing the depth to three layers does not improve upon the best shallower configuration on any of the three datasets.
Thus, POEM achieves its best performance with only one or two predictor layers in the evaluated settings.
This finding is consistent with our motivation that phase-guided feature rotation makes temporal patterns more regular and allows the subsequent forecasting task to be handled by a lightweight predictor.

\begin{table}[t]
\centering
\renewcommand{\arraystretch}{1.1}
\setlength{\tabcolsep}{3pt}
\fontsize{10}{12}\selectfont
\rmfamily

\begin{tabular}{lccc}
\toprule
\textbf{Dataset}
& \textbf{POEM (ms)}
& \textbf{PMDformer (ms)}
& \textbf{SRSNet (ms)} \\
\midrule

Weather
& 0.441
& 0.268
& 1.102 \\

ETTh2
& 0.242
& 0.169
& 0.878 \\

ETTm1
& 0.246
& 0.225
& 1.009 \\

ETTm2
& 0.201
& 0.150
& 0.870 \\

US Birth
& 0.249
& 0.240
& 0.999 \\

\bottomrule
\end{tabular}

\caption{Average inference time per sample of POEM, PMDformer, and SRSNet across different datasets. Detailed results are reported in the Appendix.}
\label{tab:efficiency}
\end{table}

\subsection{Efficiency Analysis}
We compare the inference efficiency of POEM with PMDformer and SRSNet on datasets Weather, ETTh2, ETTm1, ETTm2 and US Birth.
Table~\ref{tab:efficiency} reports the average inference time per sample over four prediction horizons.

POEM requires $0.201$--$0.441$\,ms per sample across the five datasets.
Its average latency is $0.009$--$0.173$\,ms higher than that of PMDformer, but it is approximately $2.5$--$4.3$ times faster than SRSNet.
The latency gap to PMDformer is largest on the 21-variable Weather dataset and smallest on the univariate US Birth dataset, consistent with channel-dependent overhead from the variable-wise phase correction and feature rotation.

Overall, although POEM introduces additional phase correction and feature rotation operations, its measured inference latency remains close to PMDformer and substantially lower than SRSNet.

\begin{figure}[ht]
  \begin{center}
  \includegraphics[width=\linewidth]{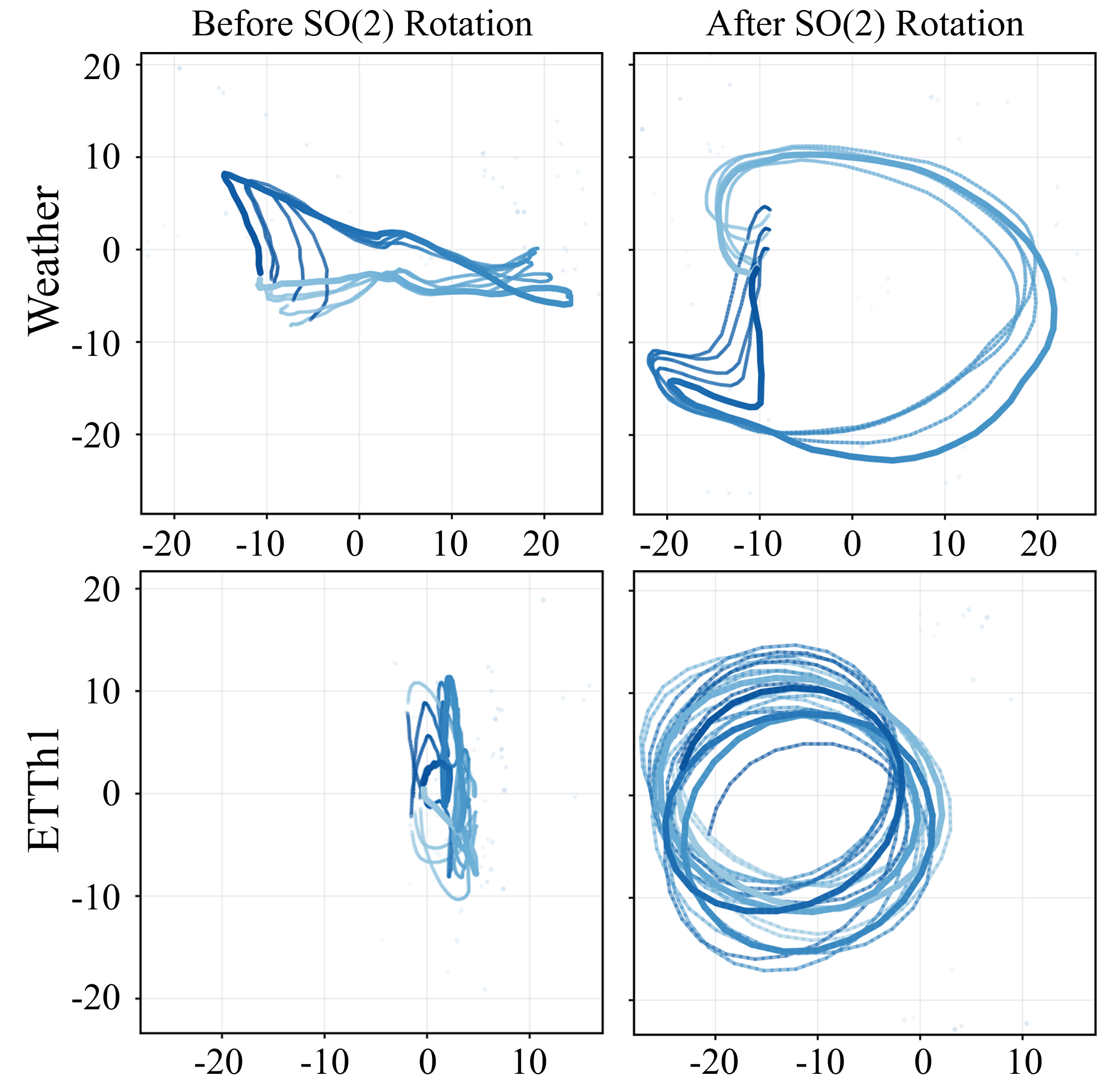}
  \caption{
    Two-dimensional PCA projections of feature trajectories before and after the phase-guided $\mathrm{SO}(2)$ rotation on Weather (up) and ETTh1 (down).
  }
  \label{fig:visualiztaion}
  \end{center}
\end{figure}

\subsection{Visualization of $\mathrm{SO}(2)$ Feature Rotation}
To qualitatively examine the effect of the phase-guided $\mathrm{SO}(2)$ feature rotation, we project the latent feature trajectories immediately before and after the learned rotation into two dimensions using PCA.
We jointly fitted a PCA basis to all the features before and after rotation.
As shown in Figure~\ref{fig:visualiztaion}, we present two cases from ETTh1 and Weather.
In the selected cases, the post-rotation trajectories are smoother and form more coherent cycles than the corresponding pre-rotation trajectories, consistent with the intended reduction of phase-related variability, qualitatively demonstrating the effectiveness of POEM in addressing periodicity drift.

\section{Conclusion}
In this paper, we present POEM, a phase-aware forecasting framework designed to mitigate periodicity drift.
POEM constructs a phase correction trajectory for each variable from local phase offsets and velocities, and applies phase-guided $\mathrm{SO}(2)$ rotations to produce more regular feature representations for a lightweight temporal predictor.
Directional Phase Increment Attention further extends the phase correction into the forecast horizon by retrieving historical phase increments observed under similar timestamp contexts.
Experiments on nine benchmarks show competitive forecasting accuracy, while ablation results support each component and PCA visualizations show that the rotation produces more regular feature trajectories.
Efficiency analysis indicates that POEM keeps its average per-sample inference time below $0.5$\,ms while substantially outperforming SRSNet.
Future work will explore cross-variable phase modeling and extend the framework to more complex periodic structures.

\bibliography{aaai2027}


\end{document}